# Visual Attack and Defense on Text

Shengjun Liu[1], Ningkang Jiang[2], Yuanbin Wu[3]

[1,2,3]East China Normal University
[1]51184501131@stu.ecnu.edu.cn, [2]nkjiang@sei.ecnu.edu.cn, [3]ybwu@cs.ecnu.edu.cn

**Abstract.** Modifying characters of a piece of text to their visual similar ones often appear in spam in order to fool inspection systems and other conditions, which we regard as a kind of adversarial attack to neural models. We propose a way of generating such visual text attack and show that the attacked text are readable by humans but mislead a neural classifier greatly. We apply a vision-based model and adversarial training to defense the attack without losing the ability to understand normal text. Our results also show that visual attack is extremely sophisticated and diverse, more work needs to be done to solve this.

## 1 Introduction

When reading a piece of text, humans need to judge the meaning of the characters from their appearance before understanding the text. In the face of characters that have not been seen before (visual text attack), we can infer the meaning of the characters if we have seen visual similar characters: 'ạpplє' means 'apple'. Most of current NLP models initialize random embeddings for characters or words based on their identities and update the embeddings during training, visual text attack may cause them to fail to understand text. Recently, some work encodes characters with visual information, (Dai and Cai, 2017) tries to apply the image features of Chinese characters to language modeling and word segmentation tasks. In the real world, there is much malicious behavior trying to utilize visual vulnerabilities of models, such as spam (Hosseini et al., 2017), dirty talk in games and forums, etc.

The research on adversarial attack has been paid much attention recently, which aims to reveal the weakness of neural models and improve robustness of the models against various attacks such as replacement and deletion of words. Compared with the general attack on NLP systems: 1) Visual attack is to replace a character **X** in the original text with another character **X'** that looks like **X** and no knowledge of language is required. 2) Humans are more robust against visual attacks than grammatical attacks (Hosseini et al., 2017). For a character X, (Eger et al., 2019) utilizes the python PIL package and a font file to draw **X** on a (length: 24, width: 24) image(Fig. 1. Left), stacks the rows of the image into a one-dimensional (24*24=576) vector as an image-based character embedding (ICE) of **X**, then calculates cosine distance with other characters based on ICEs to find the nearest neighbors as disturbances. Problem with this method is ICE neighbors are based on pixel matching of images but their two-dimensional information is not fully used, which may cause that some visually similar neighbors with different sizes and positions cannot be found. As shown in Table 1(For illustration, characters and their neighbors shown in this paper are in a font named **Dejavu Sans** used in our experiments), many ICE neighbors of character q are not like q and even difficult for humans to recognize. (Zhao et al., 2017) inverts a picture **P** with CNN into a one-dimensional hidden embedding z, disturbing z and then utilizing W-GAN to convert z to an image **P'** as a disturbed sample. For printed characters there are two problems: 1) There is only one image for every character in a font file, CNN is easy to overfit; 2) **P'** can't be directly taken as neighbors because **P'** is very likely not include in a font(models yield decimals while pixels from fonts contain only integers). If **P'** is used to find neighbors, it becomes pixel matching like ICE which is not reasonable. On defense: (Eger et al., 2019) takes ICEs as embeddings of characters and neural models do suffer less from visual text attack.

| Char | ICES | I2CES | DCES |
|---|---|---|---|
| H | H H H Ḥ Ḫ Ḩ Ḣ H̡ Π Π | H H H ʜ ʜ Ḥ Ḣ Һ ᴴ ᴴ | H Ͱ ℌ Ȟ Ⱨ Ħ Ḧ ᴴ H̡ Ĥ |
| q | q ɋ g ɡ g ǵ ġ ɑ ḡ ɖ | q ɥ ɋ ɋ ȹ α̨ ợ ɋ g g | q ʠ ɋ |
| 3 | 3 Ǝ Ǝ 8 ȝ ⅎ ⅎ 9 Ƨ S | 3 ʒ Ʒ ʒ ʒ ³ ʒ̧ ³ Ȝ ₃ | |
| 4 | ⇞ 1 ⇕ ↑ ⬍ ↕ ⬆ ⇂ ⇃ ǂ | Ч ⁴ ₄ ☆ ꝝ Ч Œ ц ɋ ɋ | |

**Table 1.** up to 10 nearest neighbors in different character embedding spaces

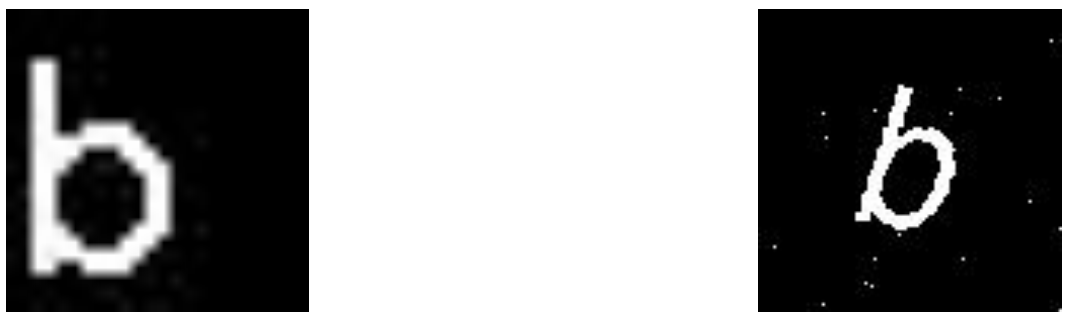

**Fig. 1.** letter 'b' drawn by PIL package: Left: (24,24) image; Right: (100, 100) image with centering, rotation and noise.

Our work on attack: We utilize CNN to get visual information of characters, augment data of characters and get more visually similar characters; on defense: we apply a vision-based neural model and adversarial training and show that they are useful for resisting visual attack.

## 2 Method

To better deal with visual text adversarial attack, we apply 1) a vision-based perturber, 2) a vision-based neural model.

### 2.1 Vision-based Text perturbations

We hope that after visual text perturber(VP) disturbs certain characters in a input text, humans can still understand the text correctly, while neural models may fail. The parameters of VP are a character **X**, a probability **p** and a character embedding space (CES). Each character has up to 20 nearest neighbors in the CES for preparation. For

| | |
|---|---|
| Conv2d (16, 5, 5, 0), MaxPool2d(2, 2), ReLu | |
| Conv2d (32, 3, 3, 1), MaxPool2d(2, 2), ReLu | |
| Conv2d (64, 3, 3, 1), MaxPool2d(2, 2), ReLu | |
| Conv2d (128, 3, 3, 1), MaxPool2d(2, 2), ReLu | |
| Conv2d (128, 3, 3, 1), MaxPool2d(2, 2) | |
| Linear(1152, 500), Linear(500, num_class) | |

**Table 2.** architecture of the model used for VP training. Parameters of Conv2d: number of output channels, kernel size, padding

each character in text there is **p** probability that it will be replaced. If a character **X** is disturbed, one element in neighbors of **X** is randomly selected. We represent VP as:

VP = VP (**X**, **p**, CES)

For CES, (Eger et al., 2019) proposes image-based character embedding space (ICES) and description-based character embedding space (DCES). ICES is based on ICE introduced in §1. DCES first gets description of a character from the Unicode 11.0.0 final names list, for example, the description of 'b' is 'LATIN SMALL LETTER B', characters whose descriptions contain the same character and the same case are considered as neighbors of 'b', such as 'ƀ', description of which is 'LATIN SMALL LETTER B WITH STROKE', these two descriptions both include 'SMALL' and 'B'.

We propose image-based 2-dimensional character embedding space(I2CES) which utilizes more features of images. The images of characters drawn with PIL are input to a CNN classifier. The architecture of the classifier is shown in Table 2, which is a slightly modified version of VGG-16(Simoyan and Zisserman, 2014). In order to capture more visual information, we expand input image size(24, 24) to (100, 100). The characters are centered(shown in Fig. 1. Right) to make the model focus on the shape and to deal with characters that look very similar but in different positions. To prevent the classifier from overfitting we augment training data through: eight different fonts(differ by thickness and whether oblique); two font sizes {60, 80} to make the model robust to different sizes; rotating the characters from -20 to 20 degrees by the center of images, the interval is 2; adding salt-and-pepper noise. The final number of each character's samples is 8*2*20=320. We suppose that when the model converges and the accuracy is more than 90%, the model has mastered visual features of the characters like human beings to a certain extent. For a character **X**, we take a specific sample of **X** as input to the trained classifier: font size 80, no rotation, a font (named Dejavu Sans) chosen from the 8 fonts mentioned before. Rather than applying output of the final linear layer, we take output of the last convolutional module and flatten the output as I2CES, reason of this decision is discussed in §4. Cosine similarity is used to calculate distance between embeddings.

### 2.2 Vision-based neural model

Many models encode text in character level and have achieved very good results, we use Char-CNN (Zhang et al, 2015) as the base model. Char-CNN encodes characters with one-hot vectors and processes them with 1-Dimension Convnets. In (Eger et al., 2019), ICEs are directly fed into the neural model as embeddings of characters. Like (Shimada et al. 2016) and (Dai and Cai 2017), we put a naive CNN (3 convolution layers and 2 MaxPool layers) module into Char-CNN which aims to utilize the visual information of characters. As illustrated in Fig. 2, for each character **X** in input text, the image of **X** is fed into a 2D CNN, the flattened features of the last convolution layer are taken as embeddings of the characters. The 2D CNN is trained together with Char-CNN.

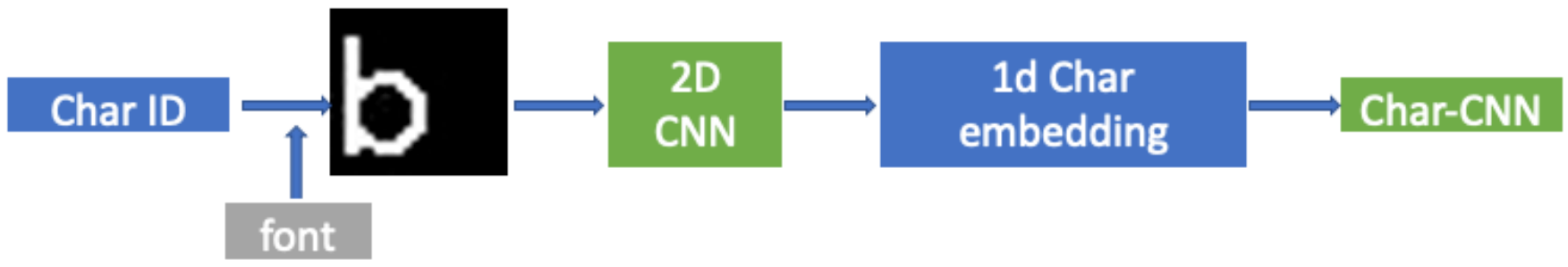

**Fig. 2.** Vision-based neural model: 2D-CNN module takes a character's image and outputs an embedding, Char-CNN module processes the embeddings of characters in text

| Char | I2CES | ***h_set*** |
|---|---|---|
| q | q ɋ ɋ ʠ ɋ ɑ̨ ợ ꝗ ɡ g | q ɋ ɋ ʠ ɋ ợ ꝗ |
| 4 | ᑫ ⁴ ₄ ☆ ϟ ᑫ Œ ц ɋ ɋ | ᑫ ⁴ ₄ ϟ ᑫ ц |

| Char | ICES | ***h_set*** |
|---|---|---|
| q | q ɋ ɡ g ɡ ǵ ġ ɑ ḡ ɖ | q ɋ |
| 4 | ⇞ 1 ↨ ↑ ⬍ ↕ ⬆ ⇂ ⇃ ǂ | |

**Table 3.** neighbors in ICES and I2CES and corresponding effective ones

## 3 Experiment

We perturb characters with I2CES and see how vanilla Char-CNN responses to the visual attack. Then we evaluate the effect of several defense methods. The experiment environment is pytorch 1.4.0 and machine is GTX1080ti 11GB GPU.

### 3.1 Datasets

We choose three classic classification datasets:
AG's news corpus: 4 classes, 120000 train and 7600 test samples,
DBPedia ontology dataset: 14 classes, 560000 train and 70000 test samples,
Yelp reviews: 5 classes, 650000 train and 50000 test samples.

### 3.2 Perturb

For I2CES, dimension of the output got from the convolutional layer(Conv2d (128, 3, 3, 1), MaxPool2d(2, 2) in Tabel 2) is (128,3,3), dimension of final embeddings of characters is (1, 1152). 4378 characters can be found to have visual neighbors, so class num-ber of the CNN in §2.1 is 4378. All these characters need to be included in the 8 fonts mentioned in §2.1 and images of which cannot be blank (Some Unicode numbers are in these fonts while their pictures drawn by PIL have no black pixels). In Table 1 we list neighbors of some characters in ICES, DCES and I2CES. We pick effective (visually similar judged by 3 volunteers) ones from ICES, I2CES (Table 3 shows some examples) and take them as new replacement set, ***h_set***. Number of ***h_set***(letters a-z & A-Z) in ICES and I2CES are 572 and 697, showing that I2CES successfully capture more visual information than ICES. Compared to ICES, neighbors found by I2CES are more diverse: some of the neighbors are different in character's positions and sizes in images. Compared to DCES, I2CES is more general: DCES is limited to commonly used English letters, numbers and many other symbols can't be perturbed by DCES, while I2CES is based on visual information, most characters (such as number '3' and '4' shown in Table 1) in fonts have the corresponding replacement set, which is applicable in more scenarios.

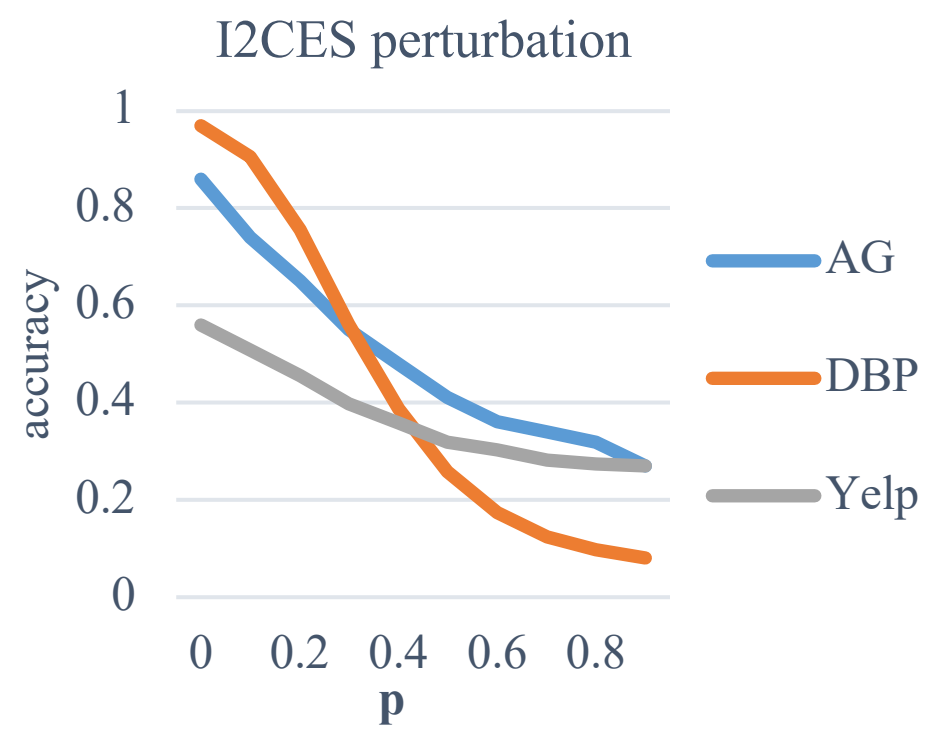

**Fig. 3.** Accuracy degradation of origin classifier under I2CES attack

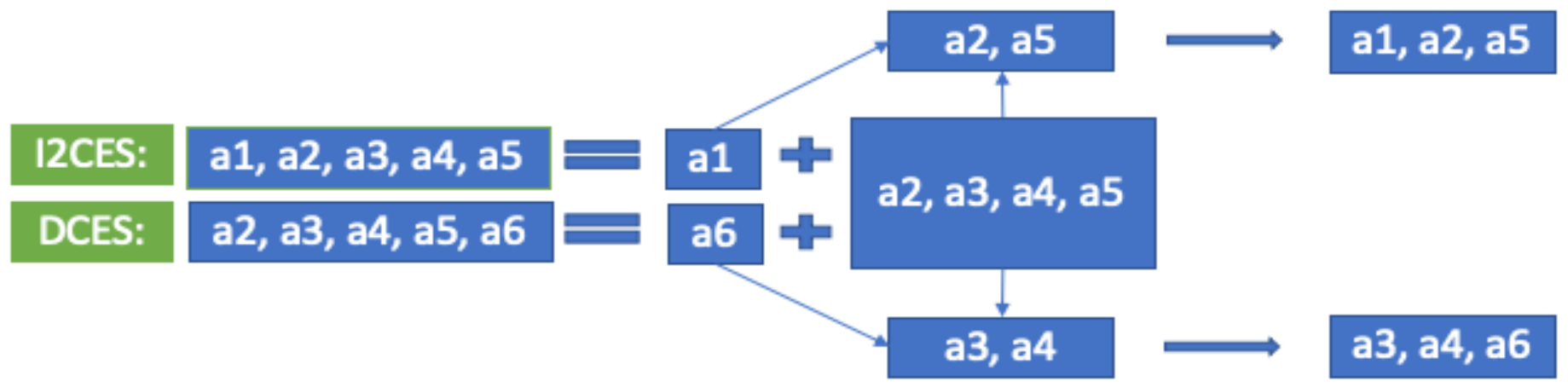

**Fig. 4.** Half of intersection of neighbors are chosen for DCES, the left half are for I2CES

We perturb test sets in the three datasets described in §3.1 with ***h_set*** in I2CES, Fig. 3 shows the great degradation of accuracy and degrades more when p increases. A character **X** not in predefined vocabulary are marked as 'unknown' and an all-zero vector is taken as the embedding of **X**, causes the model to misclassify the input.

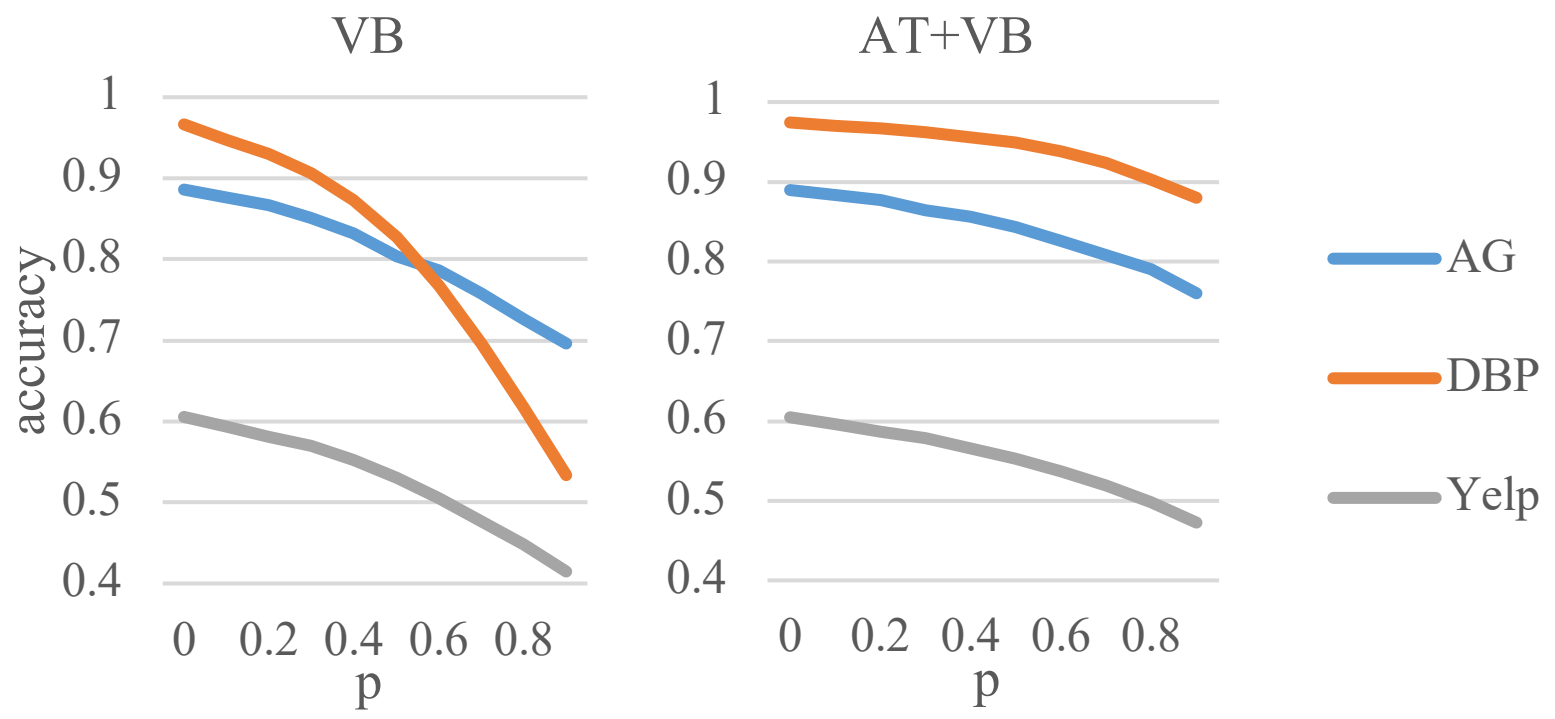

**Fig. 5.** Accuracy degradation of VB classifier under DCES attack

### 3.3 Defense

Our methods on defense: the vision-based model (VB) described in §2.2, adversarial training & vision-based model (AT+VB). For AT, we disturb train set with ***h_set*** in I2CES and train on the disturbed data after training on clean data. We test robustness of the trained model with DCES neighbors as in (Eger et al., 2019) because of its high readability by humans. In adversarial training, (Eger et al., 2019) perturbs the input in train set with all the 20 ICES neighbors, there will be two problems: 1) as mentioned in §1, there are some characters in ICES and I2CES that are not like original characters. Training a model with dissimilar characters will cause the model to be confused and cannot resist visual attack well; 2) the neighbors of ICES and DCES partially overlap, which causes part of the test set appears in the train set and has an impact on the authenticity. The improvements we propose(shown in Fig. 4): we first get intersection of ***h_set*** in I2CES and neighbors in DCES, then randomly take half of the intersection as neighbors in DCES to test robustness of the model and the left half as neighbors in I2CES for training, which aims to better check the model's ability to capture visual features. Fig. 5 draws the effect of defense: VB is more robust to visual text attack than origin Char-CNN and AT+VB can get better results. Especially for DBPedia, AT+VB gets 35% accuracy than VB, shows that adversarial training with effective adversarial examples makes the model more robust. Meanwhile, both VB and AT+VB can achieve as good performance on clean test texts as origin Char-CNN, shows that VB can capture both visual and semantic information of characters.

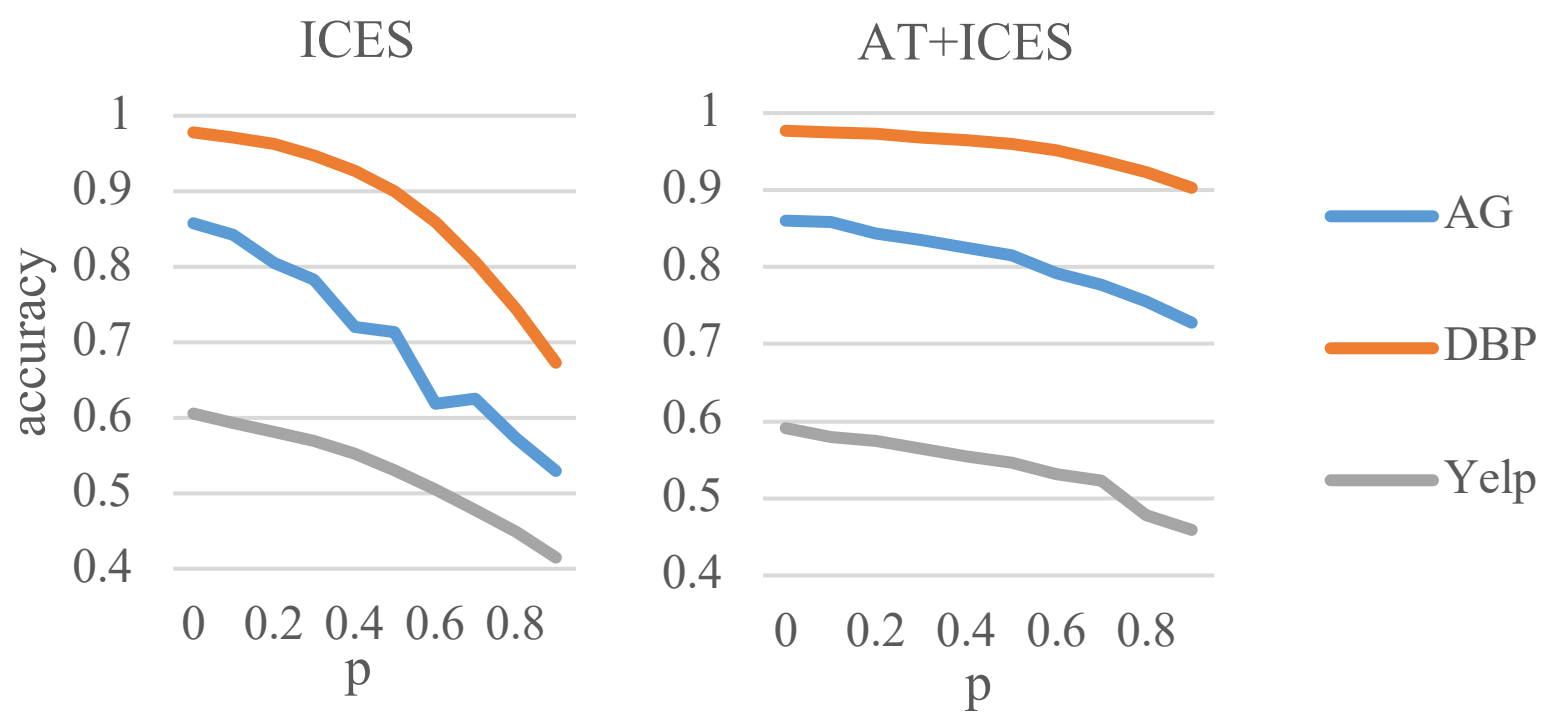

**Fig. 6.** shows the accuracy degradation on ICES based Char-CNN and AT+ICES based Char-CNN under DCES attack

| ‘b’ | origin | ***h_set*** |
|---|---|---|
| linear | ƅ ▮ Б Ƃ Б ɒ ƀ D D ➤ | ƅ ƀ |
| linear + ave | Ъ Ь h h h ɓ Ь Ϧ ƕ ʖ | Ъ Ь ɓ Ь |
| conv | Ъ ɒ Ƃ ɓ Ь ɓ ϭ ᵇ Ɒ h | Ъ Ƃ Ь ɓ ᵇ |
| conv + ave | Ъ Ђ Ь ɓ h h h ƕ Ϧ ᵇ | Ъ Ђ Ь ɓ Ϧ ᵇ |

**Table 4.** neighbors in I2CES got from different layers

We also train ICES-based Char-CNN with adversarial examples as VB, Fig. 6 shows that ICES can get comparable results with VB on these datasets. Some possible explanations: 1) we get at most 20 neighbors on CES for each character, the neighbors are too few to form a distribution. This leads to what the classifier remembers are samples rather than distribution of the samples. In VB trained by AG, cosine similarity between embeddings by the 2D-CNN module of ‘z’ and ‘ƶ’ is 0.77, and 0.69 with ‘Ζ’, which shows that the module can’t handle transformations very well. 2) (Meng et al., 2019) points out that architecture of the 2D-CNN module needs to be specifically designed to capture visual features when dealing with small character images.

Though AT+VB can achieve great improvements against visual text attack, accuracy drops up to 15% in yelp_review. Part of the drop is on account of the visually dissimilar neighbors in DCES. As shown in Table 1, ‘Ͱ’ and ‘ℌ’ are both DCES neighbors of letter ‘**H**’, it’s hard for human to recognize without linguistic knowledge.

## 4 Discussion

For I2CES, how to get more visually similar neighbors is important. We compare output from two different modules of the trained classifier described in §2.1: the last convolutional module and the final linear module(Linear(500, num_class) in Table 2), dimension of which are 1152 and 4378 respectively. Examples of character ’b’ from are illustrated in Table 4, we list 10 nearest neighbors in ‘origin’ and take ***h_set*** from these neighbors. 'ave' means taking 20 images of the 320 ones (mentioned in §2.1) of a character: a chosen font(name: Dejavu Sans), font size 80, rotation angle from -20 to 20 degrees, the interval is 2. Results from convolutional layers are much better than former one. Number of characters in ***h_set*** of **linear** is very small and **linear + ave** version is a bit higher. **Conv** version are comparable with **conv + ave** version while the **conv** versions are much better than **linear** versions respectively. We consider that linear layers aim at better classifying and get the most possible label which leads to ignoring unimportant ones, while main purpose of convolutional layers is taking visual features of images. **Conv** version is finally selected is because **conv** version costs less computation resources than **conv + ave** version while having comparable performance.

## 5 Related Work

Work on adversarial attack and vision-based models are important for our work.

**Adversarial Attack** aims to fool neural models by modifying models' input, while humans can still understand the input. In NLP, (Li et al., 2018) follows the general steps of attacking NLP systems: they calculate the importance of characters with jacobian matrix or a score function and modify the most important ones. These operations need to get information the neural models, while our method is black-box (no use of models) and preserves high readability of attacked texts for humans. Adversarial training (Goodfellow et al., 2015) is a common method: disturbing training data in the way of disturbing test data. Adversarial training is not usually very useful, such as against more complex attack (Carlini and Wagner, 2017).

**Vision-based models** were proposed by (Shimada et al., 2016) and (Dai and Cai, 2017) to cope with languages like Chinese and Japanese. They generate character embeddings with a convolutional autoencoder and show improvement on author and publisher identification tasks. (Liu et al., 2017) creates embeddings from a CNN and shows that models with these embeddings are more useful for rare characters. (Eger et al., 2019) gets embeddings of characters by flattening the character images and improves the robustness of model against visual text attack.

## 6 Conclusion and Future Work

In this work, we regard visual text attack as a kind of important adversarial attack in NLP because it's normal in the real world. To get more human-readable adversarial examples, we incorporate 2-D visual features of characters, which we call I2CES. We show that I2CES perturbations on text cause traditional neural classifiers to misunderstand meaning of the text on three classification datasets. We apply a vision-based model and results show that the vision-based model is more robust to visual attack without losing the accuracy on clean examples.

Our work can be easily applied to real systems which is helpful for the systems facing visual attack such as spam, junk messages.

Results also show that more future work needs to be done: 1) linguistic knowledge should be integrated into models to defense attack from different domains. 2) languages such as Chinese, Korean are much more sophisticated than English in glyph structure and visual attack in these languages is also frequent in real life, so attack on these languages should be paid attention.